\documentclass[lettersize,journal]{IEEEtran}
\usepackage{amsmath,amsfonts}
\usepackage{array}
\usepackage{textcomp}
\usepackage{stfloats}
\usepackage{url}
\usepackage{verbatim}
\usepackage{graphicx}
\usepackage{booktabs}
\usepackage{float}    
\usepackage{subcaption}
\usepackage{amssymb}
\usepackage{comment}
\usepackage{multirow}
\usepackage{longtable}
\usepackage{rotating}
\usepackage{algorithm}
\usepackage{algpseudocode}
\usepackage[utf8]{inputenc}
\usepackage{tikz,xcolor}
\usepackage{cite}
\usepackage[implicit=false]{hyperref}

\hypersetup{hidelinks,
	colorlinks=true,
	allcolors=black,
	pdfstartview=Fit,
	breaklinks=true}
\definecolor{lime}{HTML}{A6CE39}
\DeclareRobustCommand{\orcidicon}{
\begin{tikzpicture}
\draw[lime, fill=lime] (0,0)
circle[radius=0.16]
node[white]{{\fontfamily{qag}\selectfont \tiny \.{I}D}};
\end{tikzpicture}
\hspace{-2mm}
}
\foreach \x in {A, ..., Z}{%
\expandafter\xdef\csname orcid\x\endcsname{\noexpand\href{https://orcid.org/\csname orcidauthor\x\endcsname}{\noexpand\orcidicon}}
}

\DeclareUnicodeCharacter{FF0C}{\,}  
\begin{document}

\title{A Time-Series Data Augmentation Model through Diffusion and Transformer Integration}

\author{Yuren Zhang\hspace{-1.5mm}\orcidA{}, Zhongnan Pu\hspace{-1.5mm}\orcidB{}, Lei jing\orcidC{}~\IEEEmembership{Member,IEEE}}


\markboth{IEEE TRANSACTIONS ON HUMAN-MACHINE SYSTEMS}%
{Shell \MakeLowercase{\textit{et al.}}: A Sample Article Using IEEEtran.cls for IEEE Journals}


\maketitle

\begin{abstract}
With the development of Artificial Intelligence, numerous real-world tasks have been accomplished using technology integrated with deep learning. To achieve optimal performance, deep neural networks typically require large volumes of data for training. Although advances in data augmentation have facilitated the acquisition of vast datasets, most of this data is concentrated in domains like images and speech. However, there has been relatively less focus on augmenting time-series data. To address this gap and generate a substantial amount of time-series data, we propose a simple and effective method that combines the Diffusion and Transformer models. By utilizing an adjusted diffusion denoising model to generate a large volume of initial time-step action data, followed by employing a Transformer model to predict subsequent actions, and incorporating a weighted loss function to achieve convergence, the method demonstrates its effectiveness. Using the performance improvement of the model after applying augmented data as a benchmark, and comparing the results with those obtained without data augmentation or using traditional data augmentation methods, this approach shows its capability to produce high-quality augmented data.
\end{abstract}

\begin{IEEEkeywords}
Deep learning, Generative model, Diffusion denoising model, Transformer
\end{IEEEkeywords}

\section{Introduction}
\IEEEPARstart{W}{ith} the development of artificial intelligence (AI), numerous tasks in the real world have been accomplished through technologies combined with deep learning. Typically, a neural network that exhibits excellent performance requires a substantial amount of data for training. Various types of multi-modal data, such as images, speech, and audio, can now be easily obtained from the Internet. The acquisition of these types of data is no longer an issue.

However, due to privacy concerns, costs, and other factors, not all types of data can reach the scale of image or other types data. For instance, the data scale of rare diseases often remains relatively small\cite{mitani2020small,griggs2009clinical}. Similarly, time-series data, due to its dynamic nature over time, typically requires capturing multiple time points for a complete dataset. As a result, data acquisition is often complex and involves high operational costs. Therefore, the construction of datasets for these types of data has consistently posed challenges for researchers.

In order to address the issue of insufficient data for model training, researchers typically employ data augmentation methods to generate large amounts of data from small sample sets\cite{nakamura2024skeleton}. For example, some traditional methods involve adding noise to the original dataset to generate new data, or more advanced and powerful neural networks, such as VAE\cite{louizos2015variational}, GAN\cite{goodfellow2014generative}, and Diffusion-based generative models, are used to generate data. Sometimes，these models are combined with Transformer. Through this way, these models can capture long-distance dependencies\cite{zhou2021informer} between data by incorporating transformers.

However, even in the development of generative models, there still exists a bias towards certain types of data. Due to the larger scale of image data, generative models are typically first applied to image generation, with only the most successful models in the image domain being extended to other fields, such as time-series data. As a result, the field of time-series data generation remains relatively underdeveloped. For example, the majority of current time-series generation models still rely on architectures like VAE and GAN, with only a few approaches exploring the more advanced Diffusion models.

The Diffusion model, also known as the Denoising Diffusion Probabilistic Model(DDPM) \cite{ho2020denoising,sohl2015deep}, has obtained significant attention in recent years due to its superior generative quality and its ability to circumvent the challenges associated with tuning Generative Adversarial Networks (GANs)\cite{goodfellow2014generative}. This model has achieved remarkable success in the domain of image generation, with Diffusion-based generative models such as DALL-E\cite{ramesh2021zero} and Stable Diffusion\cite{rombach2022high}, or other fields like text generation\cite{li2022diffusion} or audio generation\cite{kong2020diffwave}. The Diffusion-based model are capable of producing excellent data that closely resemble those created by humans. However, its application in the field of time-series data remains relatively under explored. 

There already has lots of methods that can generate time-series data. For example, the model proposed by \cite{esteban2017real} is using Recurrent GAN(RGAN) and Recurrent Conditional GAN (RCGAN) to generate time-series data. Both RNN and GAN are complicated to train and the RNN exists the problem of gradient vanishing during long distance training. It's very difficult to get an effective model. The Diffusion-TS model proposed by \cite{yuan2024diffusion} also uses Diffusion and Transformer. However, it also combines Fourier Transforms and Seasonal-Trend Decomposition techniques\cite{cleveland1990stl}. The combination of these key components makes the model logically complex, making it difficult to adjust according to the needs of different tasks.

In summary, we have designed a time-series data generation model that combines Diffusion and Transformer architectures. The Diffusion model is used to generate data for the first time step, while the Transformer predicts subsequent time steps. By leveraging the strengths of both models, we address the challenges of GAN convergence issues and the difficulty of capturing long-term dependencies in RNNs. Additionally, our model is more logically streamlined, consisting merely of a linear combination of the Diffusion and Transformer models. This simplicity ensures that anyone with a basic understanding of these models can apply it to other time-series data generation tasks by adjusting model configurations and hyperparameters.

To evaluate the performance of our model, we designed a data glove equipped with pressure sensors to collect sign language data and applied data augmentation to the collected sign language data. In the experiments, we found that due to the characteristic of the MSE loss function, which tends to minimize the global error, the generated data often converges to the mean of the target values. To address the issues caused by the MSE loss function, we designed a weighted loss function based on the characteristics of the dataset. Additionally, to make the generated data more similar to the real data, we used a method of alternating between two loss functions during training, allowing the model's generated values to better converge to the target values. 

In summary, our contribution can be summarized as follows:

\begin{itemize}
    \item We propose a model that combines Diffusion and Transformer architectures, leveraging time-domain data to generate large amounts of high-quality data from small sample sets. 
    \item We improved the training method by using a weighted loss function and an alternating training approach with two different loss functions to address the issue of the Transformer failing to converge correctly during training.
    \item We can confirm that the data generated by this model exhibits variation curves that closely resemble those of real data, thereby effectively enhancing the model's classification performance. This similarity ensures that the synthetic data can be used as a valuable substitute in training, improving the model's ability to generalize to unseen data.
\end{itemize}

The structure of this paper is as follows: In Section 2, we provide an overview of related work on time-series data augmentation. Section 3 presents a detailed description of the proposed model architecture. Sections 4 and 5 focus on the experiments conducted using the model and the corresponding results. Finally, in Section 6, we summarize the key findings and draw the overall conclusions.

\section{Related Work on Time-Series Data Augmentation}
\subsection{Fundamental Traditional Techniques}
\begin{figure}[!t] 
    \centering
    \includegraphics[width=0.5\textwidth]{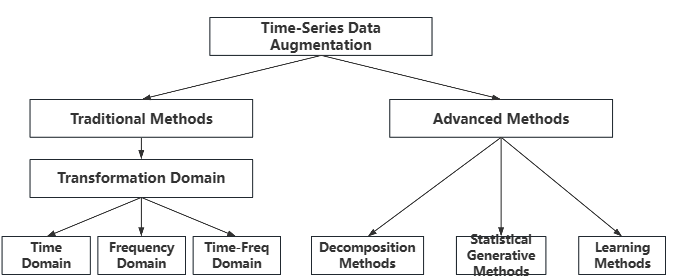} 
    \caption{Analysis domain in signal processing} 
    \label{taxonomy} 
\end{figure}

The basic traditional methods involve augment data in different domains, such as time domain, frequency domain and time-frequency domain.Just like the figure \ref{taxonomy} shows.

Data augmentation in the time domain involves processing the original data. Several methods originally developed for the image domain can also be applied to time-series data, including window cropping\cite{le2016data}, window warping, and flipping. These straightforward techniques allow us to capture essential local information, which can then be leveraged to generate additional data. Another direct approach is to introduce noise into the original data. Adding noise is a common practice across various fields to enhance the robustness and diversity of models and datasets\cite{maharana2022review}. When applied to time-series data, noise can help create new data distributions. Moreover, by varying the types of noise introduced, we can obtain different new datasets\cite{wen2020time}.


Researchers not only process data in the time domain but also transform it into the frequency or time-frequency domain using Fourier transforms. In these domains, techniques such as cropping, warping, and noise addition—including adjusting signal amplitude and phase—are used to generate new signals\cite{wen2020time}. Methods like STFT and wavelet transforms\cite{rhif2019wavelet} are especially effective for capturing non-stationary signals, and the inverse Fourier transform converts processed data back into the time domain. Frequency domain data is typically more stable and better represents the signal's periodicity and regularity\cite{cochran1967fast}.

\subsection{Advanced Methods}

\subsubsection{Decomposition}

Beyond these traditional methods, advanced approaches have been developed to leverage the unique characteristics of time-series data, aiming to capture the underlying structure of the data and generate additional realistic samples\cite{wen2020time}. For example, one decomposition method that be widely used in many fields is Seasonal-Trend decomposition using Loess(STL)\cite{cleveland1990stl}. This method decomposes the time-series data into three parts-seasonal,trend and residual.

Trend indicates the long-term direction of the time series, capturing the fundamental movement within the data. Seasonal represents the periodic contents in the signal, which is associated with specific time periods. Usually, there may be not only one seasonal component present in the time series. Combining multiple seasonal components can better capture the temporal dynamics and complex dependencies within time series data\cite{verbesselt2010detecting}. And the residual component represents the difference between the original data and the trend and seasonal components, capturing random fluctuations or noise. Once the data has been decomposed into trend, seasonal, and residual components, data augmentation can be achieved by manipulating these three components. For example, the trend component can be smoothed or altered using linear or nonlinear changes to represent different long-term trends. The seasonal component can be increased, decreased, or adjusted to produce diverse periodic variations. Furthermore, the residual component can be modified by adding various types of noise to enrich the diversity of the data.

\subsubsection{Statistical Generative Model}

Statistical generative models are typically based on explicit probability models, which assume that data follows a certain structure or distribution and model the data through a specific function to generate new data\cite{salakhutdinov2015learning}. Notable examples include Gaussian Mixture Models(GMM), Hidden Markov Models(HMM), and Autoregressive Models(AR). However, the performance of data generated by explicit probability models tends to be relatively poor because most real-world data do not adhere to a fixed structure and cannot be modeled by a specific explicit probability function. Using simple assumptions may lead to the loss of many important dependencies in the data. 

A premise of the model design in this paper is based on an important assumption of autoregressive models: the value at the first time step influences the readings at all subsequent time steps\cite{kingma2013auto,box2015time}. By adding perturbations to the data at the first time step, we can generate more new data.

\subsubsection{Learning Model}

Time-series data augmentation methods that rely on learning typically involve neural networks. By using techniques such as embeddings\cite{devries2017dataset}, reinforcement learning\cite{kaelbling1996reinforcement}, and transfer learning\cite{weiss2016survey}, these deep models can autonomously learn the underlying structures of the data. This approach alleviates the burden on researchers to manually consider data distribution. Such methods have been shown to be among the most effective and advanced in the field.

Another application of these methods is the deep generative model. Currently, there are many models in statistical generative modeling, such as Variational Autoencoders(VAE)\cite{louizos2015variational} and Generative Adversarial Networks(GANs)\cite{goodfellow2014generative}, which have demonstrated excellent performance in fields like image and speech processing. The denoising diffusion model (Diffusion) used in this paper also belongs to the category of statistical generative models. But compared to traditional Diffusion models that are used in the field of image generation. We make some changes on the Diffusion model to adjust it to have ability to generate generate time-series data. The structure of Diffusion will be detailed in Section 3.

The Transformer model used in the paper is an autoregressive model, making predictions based on learned autoregressive methods in generation tasks. In fact, it can also be considered a type of deep generative model, particularly in the application to generation tasks.

\section{Methods}
\begin{figure}[!t] 
    \centering
    \includegraphics[width=2.5in]{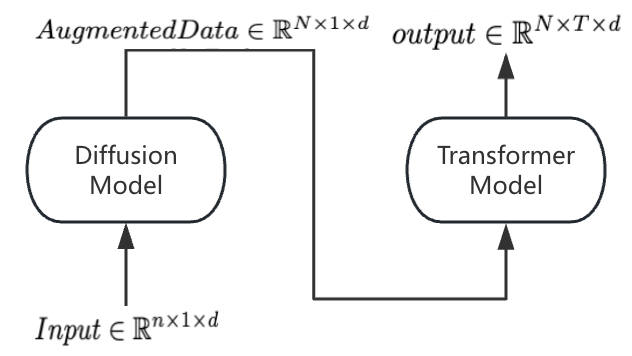} 
    \caption{Flowchart of Generative Model} 
    \label{process}        
\end{figure}

As shown in Figure \ref{process}, this is the flowchart of the model used in this paper, with the main components being the Diffusion module and the Transformer module.The input is a three-dimensional vector with a shape of \(n \times 1 \times d\), where n denotes the number of samples in the training dataset, the number 1 represents the data corresponding to the initial time step, and d indicates the dimension of the feature vector. The Diffusion model is used to generate the initial time-step data for an arbitrary number N. These data are then fed into the Transformer model, which iteratively predicts the data for the next time step until the desired input sequence length is achieved.
\subsection{Diffusion Model}

\begin{figure}[!t] 
    \centering
    \includegraphics[width=0.5\textwidth]{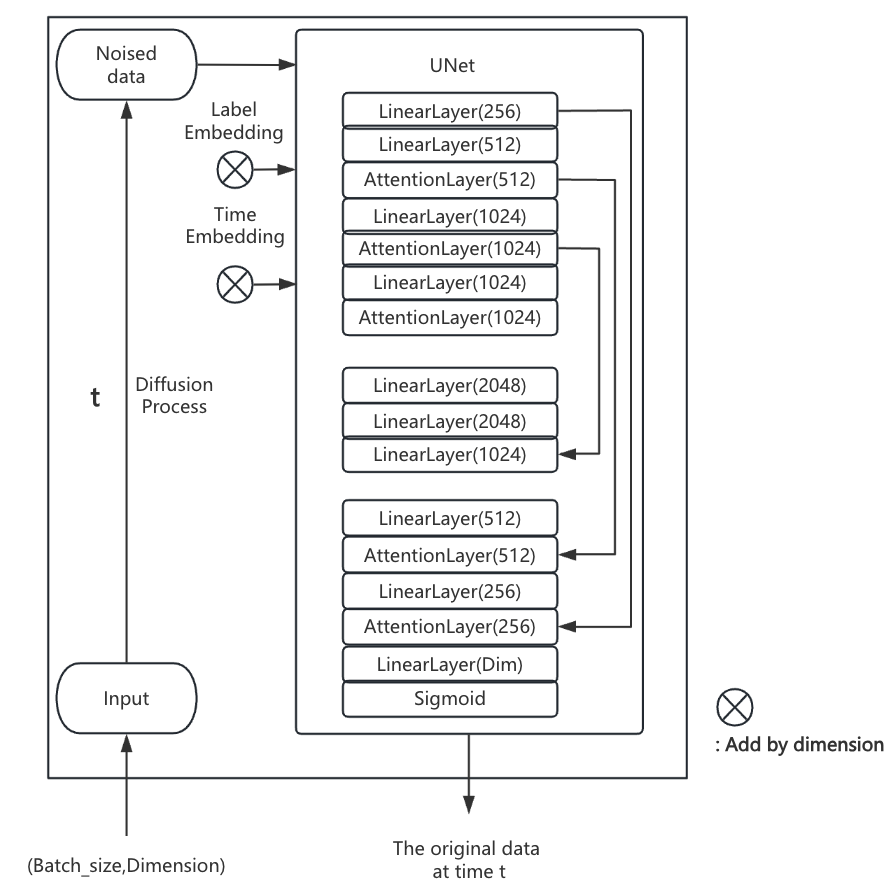} 
    \caption{Diffusion Structure} 
    \label{diffusion}        
\end{figure}
Because we followed the baseline theory that once the data at the first time step changes, the data at subsequent time steps will also change accordingly\cite{wen2020time}. Therefore, we firstly need to generate as much as possible first moment data. The Diffusion part in our model is used to generate the first moment data.

The Diffusion's structure in our model is just like the Figure \ref{diffusion}. We use the Stable Diffusion model\cite{rombach2022high}, which has achieved significant success in the field of image generation, as our foundation. And according to the needs of time-series generation, make some modifications to it.

The Diffusion model can be roughly divided into three components: input and output, the forward noise addition process, and the reverse denoising process. After the first time-step data of all samples are input into the Diffusion model, the forward noise addition process is applied, where random Gaussian noise is added to the original data. The noisy data is then fed into a UNet network for reverse denoising, where the UNet learns to restore the original data from the noisy input. After the predicted original data is output, the loss is computed by comparing it with the input, and the model parameters are adjusted accordingly. Finally, the trained UNet is used to generate new data.

We define the model's input as \( I \), where each sample \( I^i \in \mathbb{R}^d \), with \( d \) being the dimension of the features. Consequently, the entire input \( I \in \mathbb{R}^{N \times d} \), where \( N \) is the number of samples in the training set.
The entire input will be input into the Diffusion model directly without any embedding layer.

Although we use the raw data directly as input to the model, it is still necessary to normalize or standardize the data to reduce its range and lower the computational cost. We applied three methods to scale the data: [0, 1] normalization, [-1, 1] normalization, and standardization to zero mean and unit variance. To evaluate the quality of the results, we use FID as criterion to quantify the similarity between the generated data and the real data. By comparing the FID scores of the generated data, we found that [0, 1] normalization produced the best results. Therefore, we adopted [0, 1] normalization. The results will be presented in Section 4.

Once the data is input into the model, it first undergoes a forward diffusion process. A random time step t is selected, which is between[0 , T]. T is the total number of diffusion step. And standard Gaussian Noise is added to the original data according to the following formula:
\begin{equation}
I^i_t = \sqrt{\hat{\alpha}_t} I^i_{t-1} + \sqrt{1 - \hat{\alpha}_t} \cdot \epsilon
\end{equation}
Thus we can get data combined with noise. \({\epsilon}\) represents the noise. \({\hat{\alpha}_t}\) is a hyperparameter of diffusion model at time step t, which is responsible for adjusting the intensity weights between the original data and the noise.

Then input this noisy data combined with label embedding and time embedding into a U-Net network. The conditional probability distribution for reverse denoising is:
\begin{equation}
    p(I^i_{t-1} \mid I^i_t) = \mathcal{N} \left( \mu_\theta(I^i_t, t), \Sigma \right)
\end{equation}
The role of U-Net network is to learn how to predict $\mu_\theta(x_t, t)$, which represents the mean of the data at time t-1 predicted based on the noisy data at time t and time embedding. Through continuous iterations of this process, it is ultimately possible to recover the original data that is completely free of noise. Typically, U-Net is used to predict noise, because predicting the noise in an image is generally less difficult than directly predicting the original image. However, due to the smaller dimensions of time-series data, and the fact that directly predicting the original data helps us ensure the data range stays within bounds, using U-Net to directly predict the original data is more convenient for our model. The optimization objective of the model is the following formula:
\begin{equation}
\mathcal{L}(\theta) = \mathbb{E}_{t, I^i_0} \left[ \| I^i_0 - \hat{I^i_0}(I^i_t, t) \|^2 \right]
\end{equation}
\( \theta\) represents the model's parameter. The function measures the similarity between the predicted data and the real data. Additionally, we can demonstrate that directly predicting the original data yields better results. The results will be presented in Section 4.

Another change is the structure of the model. Stable Diffusion is used to generate image data, so the backbone is composed of convolutional layer. In order to generate time-series data, we use linear layer to substitute convoltional layer. The final layer of U-Net is a Sigmoid function. This function we can scale the data to be between [0, 1]. Through these processes we can make the U-Net network learn how to separate the original data from the noisy data.

\subsection{Transformer Model}

\begin{figure}[!t] 
    \centering
    \includegraphics[width=0.5\textwidth]{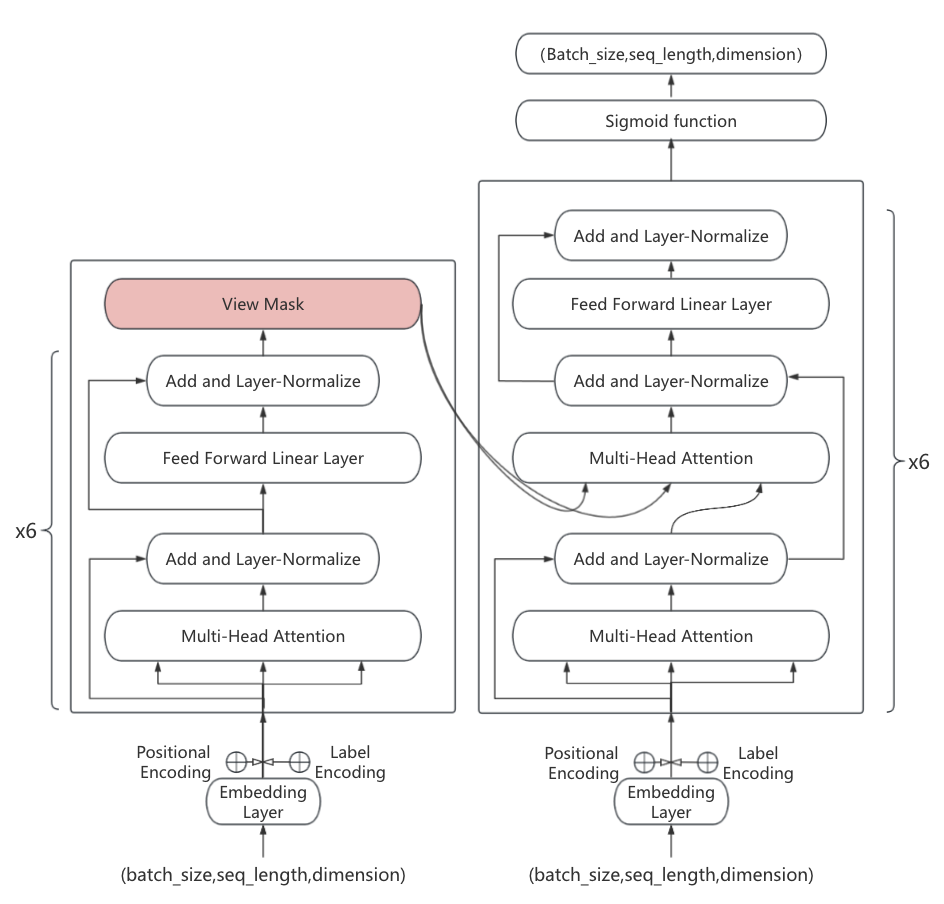} 
    \caption{Transformer Structure} 
    \label{transformer}        
\end{figure}

The Transformer is used to predict the data after the first time step. Although termed "\emph{prediction}", it actually involves generating the most likely data readings based on the distribution of the existing real data, given the data at the first time step and the sign language category. The Transformer structure used in this study is based on the traditional Transformer model, with several additional components added to achieve the prediction functionality. The overall structure is shown in Figure \ref{transformer}, including Encoder, Decoder and Loss Function as three parts. And the following sections will introduce each module of the Transformer structure used in this paper.

\subsubsection{Encoder}

To begin with, we would to introduce the input format of the encoder during training process. Before being fed into the model, the raw data is mapped to a high-dimensional space via an embedding layer. The input of the embedding layer was denoted by \( X \in \mathbb{R}^{n \times T \times d} \), where n is the number of samples, T is the sequence length minus 1 and d is the dimension of feature. As the encoder's primary role within the Transformer model is to assist the decoder in predicting the readings at specific time steps, it utilizes all time steps except the final one as reference inputs. This is precisely why the sequence length is reduced by one, which is to remove the data from the final time step. Then the raw data is first transformed into a 512-dimensional(or even higher dimension) representation through the embedding layer. Subsequently, this representation is combined with positional and label information via linear addition, forming the final input to the encoder.

\begin{figure}[!t]
    \centering
    \includegraphics[width=0.5\textwidth]{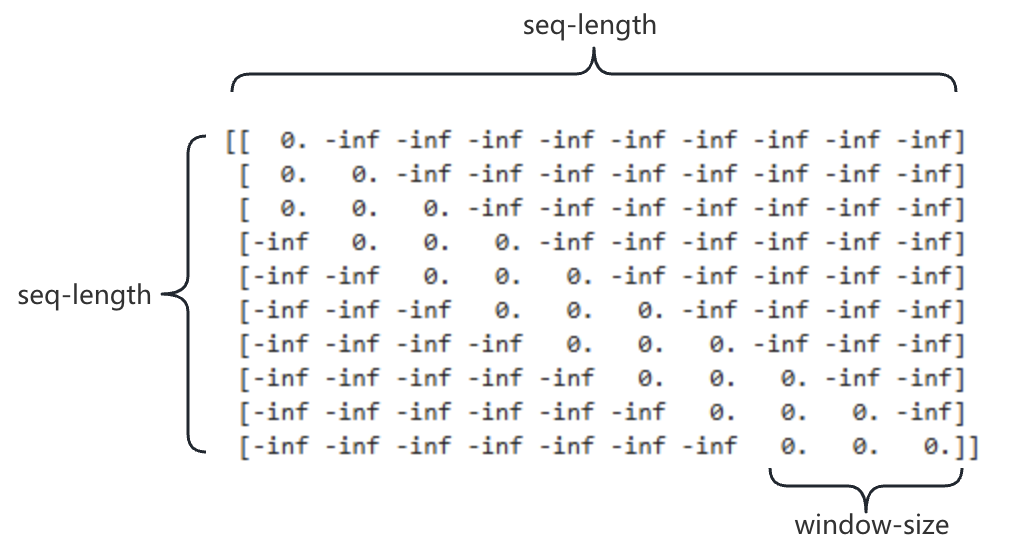} 
    \caption{View Mask}
    \label{view_mask}
\end{figure}

The backbone of encoder is the same as traditional Transformer model, composed of a multi-head attention layer, add and layer normalization and a feed forward linear layer. And this module is repeated 6 times. 

\paragraph*{View Mask}
At the end of encoder, we add a view mask. The \( \textbf{view mask}\in \mathbb{R}^{T \times T } \). It is a two-dimensional matrix, with both the horizontal and vertical axes representing the sequence length. The view mask is used to mask out data from positions that do not need attention when predicting a specific position. It will sets the self-attention weights of the previous n positions at that time step to 0, in order to adjust the predicted results. Although in the decoder the model can see all prior data up to time step t, the reference data provided by the encoder allows the model to place a significantly higher weight on the designated position to predict the data at that time step. Therefore, the view mask in the encoder plays a crucial role in the model’s predictions at each time step.

In order to conveniently adjust the position which prediction refers to, we use a hyperparameter to control the change of view mask. As shown in Figure \ref{view_mask}. The figure is "windows size=3". It means that the prediction for data at a specific time step will only be based on the data from the three most recent time steps. In this way, the reference position for the prediction can be conveniently adjusted by changing the window size hyper-parameter.

\subsubsection{Decoder}

The input to the decoder is the same as the encoder, which is a 512-dimensional vector obtained by embedding the data into a high-dimensional space and then adding position and label information. Through the decoder, we aim for it to output data with the same shape as the input, representing predictions for the next time step. The sigmoid function before the output is used to constrain the data within the [0, 1] range, keeping it within a valid range to facilitate subsequent inverse normalization back to the original data range. During training, all data except for the first time step is used as label values. The loss function is applied by comparing the decoder's predictions with the label values, and the loss is computed and backpropagated accordingly.

Another important component is the causal mask, which serves the same purpose as the mask used in traditional text sequence processing. Its role is to prevent the model from seeing data from future time steps at a given moment. In fact, the same type of mask also exists in the encoder.

\subsubsection{Loss Function}

In the early stages of model training, we use the MSE loss function. As training progresses, although the loss decreases normally, all types of time-series data generated by the model show almost no change, and the plotted variation curve appears as a straight line, making it unusable as effective augmented data. After several researches, it was found that this phenomenon occurs because during the initial data collection, in order to unify the duration of all actions, some actions with shorter durations time were intentionally extended to a longer length, with the padded part being the data from the last actual time step. As a result, many actions contain more or less a "no-change" phase. The MSE loss function calculates the deviation of each time step’s data with the same weight, and the "no-change" phase in each action causes the training direction to deviate. The model tends to predict the data to be the same as the first time step's value. Therefore, while the loss continuously decreases, the generated data shows almost no variation.

To address this question, we decided to use a weighted loss to calculate the loss between predicted data and real result. 

\begin{figure}[!t]
    \centering
    \subfloat[\textnormal{Average Feature Change Over Time}]{
        \includegraphics[width=0.4\textwidth]{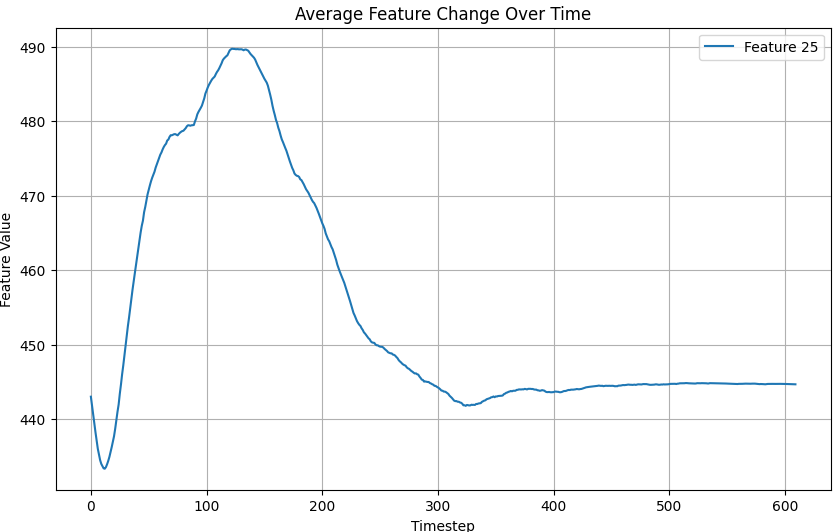}
        \label{change}
    }
    \hfil
    \subfloat[\textnormal{Difference Value Over Time}]{
        \includegraphics[width=0.4\textwidth]{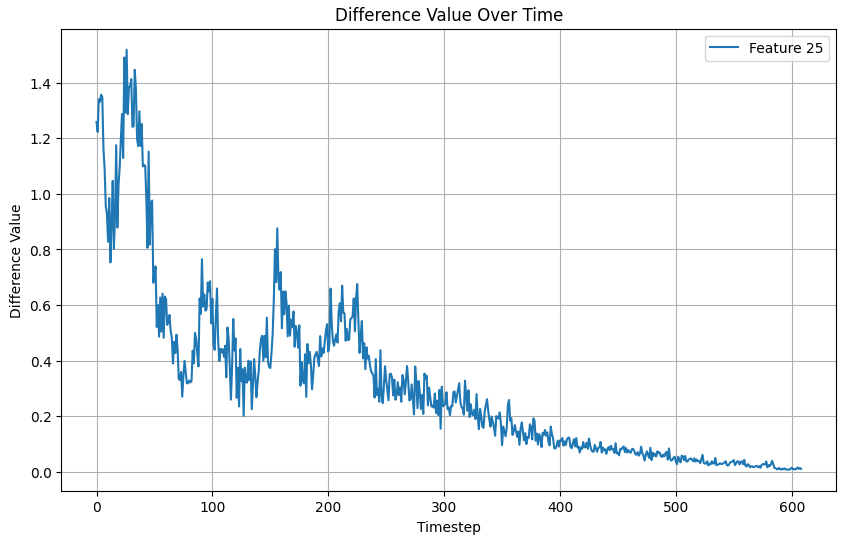}
        \label{Difference}
    }
    \vskip\baselineskip
    \subfloat[\textnormal{Fitting Curve}]{
        \includegraphics[width=0.4\textwidth]{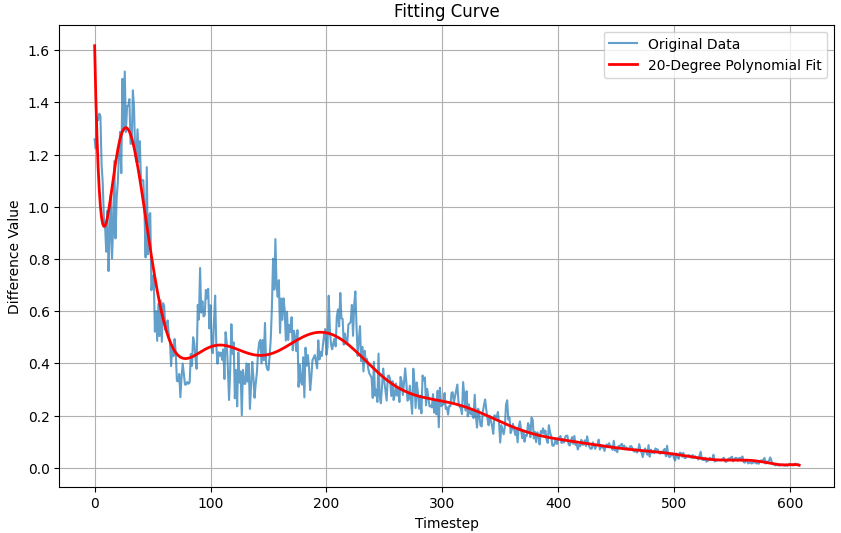}
        \label{fitting curve}
    }
    \hfil
    \subfloat[\textnormal{Quartiles of Difference Value}]{
        \includegraphics[width=0.4\textwidth]{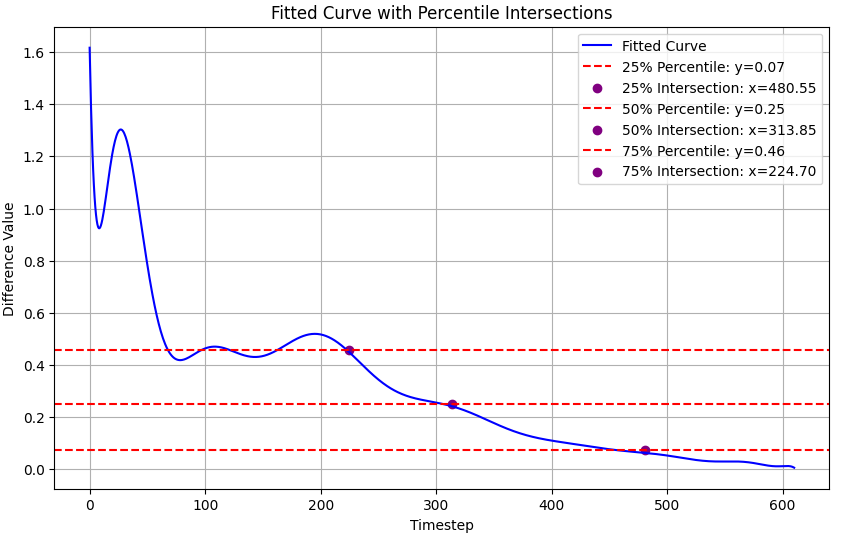}
        \label{quartiles}
    }
    \caption{The division of intervals.}
    \label{quartile}
\end{figure}

First, we define \( x_i \in \mathbb{R}^T \), \(1 \leq i \leq T \) as average feature change over time. Just like Figure \ref{change} shows. It averages the features across all samples and all dimensions during training, demonstrating the average readings of features over the time axis. The fundamental goal of performing this operation is to divide the entire long time series into shorter time segments. Although directly averaging across different action categories and features across dimensions introduces some errors and may not capture the unique transformations of each action category or each feature, it is sufficient for segmenting intervals.

To better highlight the changes at each time step relative to the previous time step, we calculate the difference 
\begin{equation}
    \Delta x_t = x_t - x_{t-1}, \quad t = 2, 3, \dots, T
\end{equation}
by subtracting the data at the previous time step from that at the subsequent time step, resulting in a difference plot between adjacent time steps, as shown in Figure \ref{Difference}.

As shown in Figure \ref{fitting curve} and Figure \ref{quartiles}, after obtaining the difference plot between adjacent time steps, a single curve is used to fit the difference plot. The purpose of this implement is to reduce the difficulty of calculation. Lots of fluctuations in the function graph may induce the question of blurred boundaries. Due to the tolerance of errors over the selected interval, the fitting curve does not need to be overly precise. In this study, a 20-degree polynomial curve was used for fitting the plot. Once the fitting curve is obtained, quartiles $Q_1$,$Q_2$,$Q_3$ are calculated based on this curve:
\begin{align}
    Q_k = \text{quartile}(\{\Delta x\}^T_{t=2}, q) \: k \in \{ 1,2,3\},q \in \{0.25,0.5,0.75\}
\end{align}
And the calculated quartile values based on the changes typically correspond to multiple time steps. To ensure that the model incorporates as many valid actions as possible, we select the rightmost time step—corresponding to the last occurrence in the temporal dimension—as the boundary between intervals. This approach results in the division of the entire time sequence into four distinct intervals.

After dividing the data into four intervals, weights are assigned to each interval accordingly. Given that the data is padded from later time steps toward the same length along the temporal dimension, earlier time steps are more likely to contain valid actions. As a result, the first interval is assigned the highest weight, while the weights for subsequent intervals decrease progressively. The loss function is defined as follows:
\begin{equation}
L = \sum_{i=1}^4 w_i \cdot  \sum_{t \in I_i} \frac{1}{d}\sum_{j=1}^d (y_{t,j} - \hat{y}_{t,j})^2,
\end{equation}
$w_i$ represent the weight for each interval, $I_i$ represent different intervals, and $y_{t,j}$ denote the specific value of the $d$-th dimension at time $t$.

Since the proposed loss function assigns higher weights to earlier time steps, the loss tends to decrease more slowly for later time steps during training. Although later time steps contain fewer valid actions, to prevent their data from adversely affecting the quality of generated data, we employed the unmodified original MSE loss function for training over shorter time segments. This approach ensures that the loss for later time steps can decrease rapidly, as equal weights are assigned at each time step. Alternating between the weighted loss and the standard MSE loss during training allows the model to achieve better predictive performance across all time steps.

\section{Experiments and Results}

To validate the quality of the data generated by the model, this study utilized an independently collected dataset of sign language data as a reference for evaluation. The data from the first moment was input into the Diffusion model after training, and the FID was used as a metric to assess the quality of the generated outputs. To ensure that the generated data closely aligned with the distribution of real data, the normalization methods and the output format of the UNet network were further adjusted, enabling a comprehensive comparative analysis of the results.

For the Transformer model, during the training phase, random Gaussian noise was added to the data to enhance the model's robustness. Additionally, two loss functions were alternated to smooth the overall loss curve of the time series. Given that the FID metric cannot effectively capture the temporal dependencies of long time series, an indirect evaluation approach was employed. First, a classifier trained on real data was used to classify the generated data, observing whether the generated data exhibited waveform variations similar to those of the real data. Subsequently, the performance of the generated data was further assessed by measuring the improvement in classification model performance when augmented data was used.

\subsection{Data Glove}
\begin{figure}[!t]
    \centering
    \subfloat[\textnormal{Data Glove Design Diagram}]{
        \includegraphics[width=0.2\textwidth]{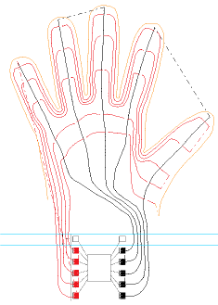} 
        \label{glove-design}
    }
    \hfil
    \subfloat[\textnormal{Key Point of Pressure Sensor}]{
        \includegraphics[width=0.2\textwidth]{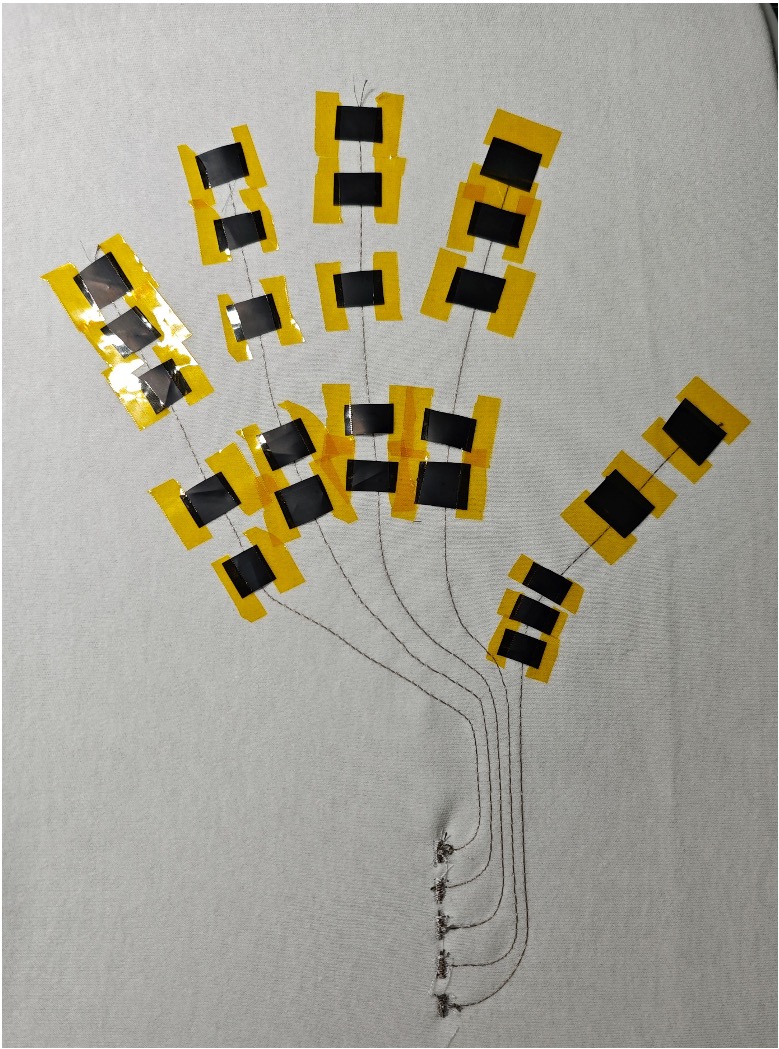} 
        \label{key point}
    }
    \hfil
    \subfloat[\textnormal{Sewn Glove}]{
        \includegraphics[width=0.2\textwidth]{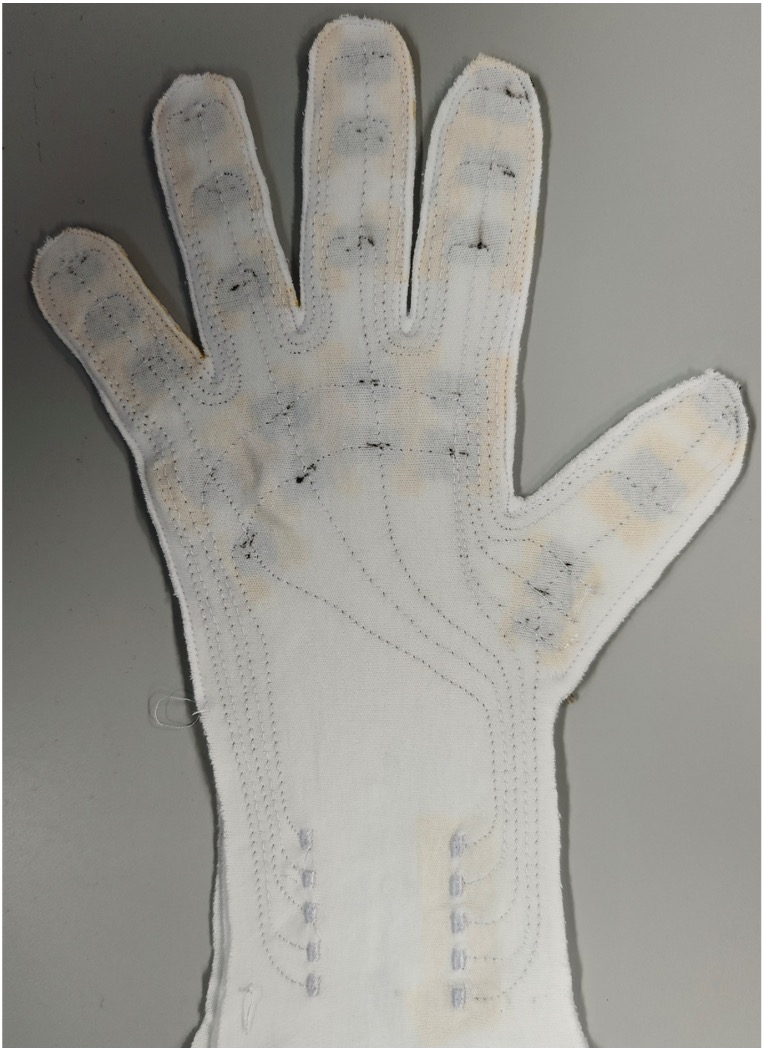} 
        \label{glove}
    }
    \hfil
    \subfloat[\textnormal{Data Glove Physical Device}]{
        \includegraphics[width=0.2\textwidth]{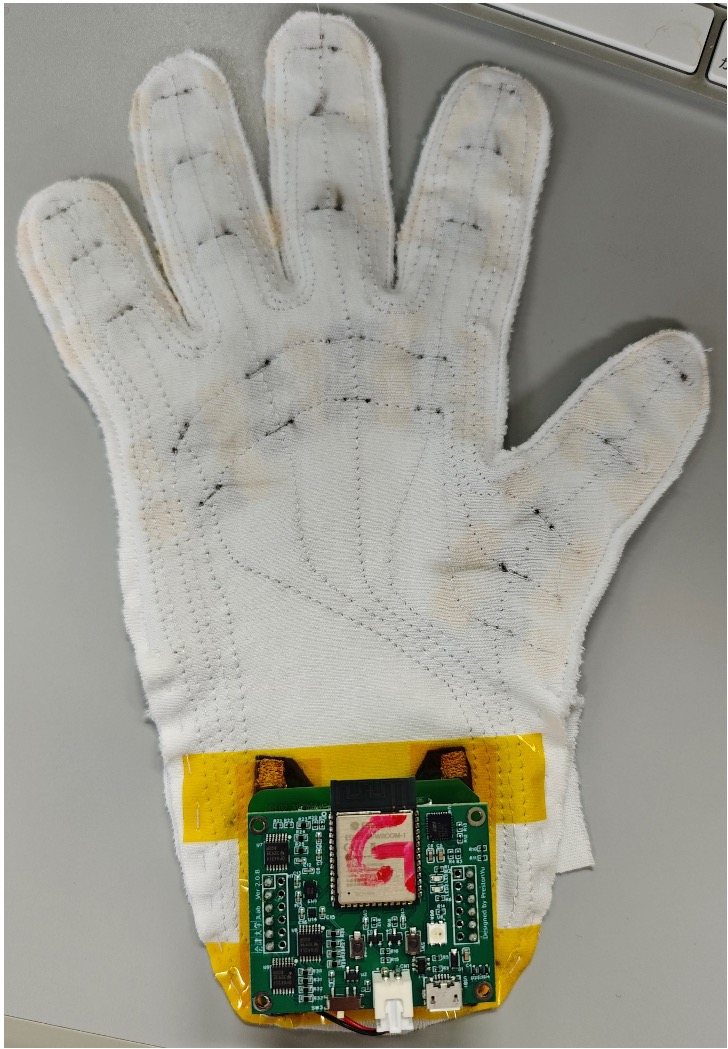} 
        \label{glove-real}
    }

    \caption{Design and Fabrication of the Data Glove}
    \label{dataglove}
\end{figure}

To collect sign language data and construct a sign language dataset, this study designed a tactile glove capable of capturing hand movements. The design of the tactile glove is shown in Figure \ref{glove-design}. The glove incorporates a total of 25 pressure-sensitive sensors made from ”Velostat”, a force-sensitive film and conductive threads\cite{kagami2024sign}. Since the changes occur primarily at the finger joints during sign language movements, the sensors are intentionally placed at each joint to accurately detect the finger movements\cite{kagami2024sign,tashakori2024capturing}. The location of key point is shown in Figure \ref{key point}. And The sewn glove is shown in Figure \ref{glove}. In addition to the pressure sensors, there is an external controller that processes the sensor data and transmits it to the terminal via a wireless network for data collection. The controller also integrates an IMU sensor to collect information such as acceleration, angular velocity, and orientation. The glove with the assembled controller is shown in Figure \ref{glove-real}.
\subsection{Dataset}

In addition to the data from the 25 pressure sensors, the IMU also includes 3-dimensional accelerometer data, 3-dimensional magnetometer data, and 3-dimensional gyroscope data, forming a 34-dimensional feature vector. The data was divided into two parts, with each part performed by five different individuals independently. We selected ten signs in total, including six dynamic signs \emph{("Best", "J", "See you later", "Single", "Yes", "Z")} and four static signs \emph{("R", "V", "W", "Y")}. Each tester performed each sign five times. For convenience in classification, all sign language actions were extended to 610 time steps, with an interval of 10ms between each steps. This resulted in two datasets, each containing 250 sign actions, which shape is (610 , 34). 

To validate the improvement of data augmentation on sign language classification, we trained a classification model using the first dataset for training and the second dataset for testing. The data augmentation model was also trained with the first dataset to generate new augmented data. We then used the combined first dataset and augmented data to train the classification model and tested it on the second dataset to observe the improvement in classification performance.

\subsection{Experiment of Diffusion Model}

As shown in Table \ref{table_comparison}, the result of generating data is as follows.
\begin{table*}[ht] 
\begin{center}
\caption{Comparison of Different Normalization and Prediction Methods on FID Scores} 
\label{table_comparison} 
\begin{tabular}{| c | c | c | c | c |}
\hline
\textbf{Method} & \textbf{Denoising Time Steps} & \textbf{Real Data} & \textbf{Generated Data} & \textbf{Avg. FID Over 10 Classes} \\ 
\hline
{[0,1] Normalization + U-Net Predict Noise} & 10 & 250 & $100 \times 10$ classes & 135.1778 \\
\hline
{[0,1] Normalization + U-Net Predict Vector (Real Result)} & 10 & 250 & $100 \times 10$ classes & 4.2511 \\
\hline
{[-1,1] Normalization + U-Net Predict Noise} & 10 & 250 & $100 \times 10$ classes & 27.8255 \\
\hline
{[-1,1] Normalization + U-Net Predict Vector (Real Result)} & 10 & 250 & $100 \times 10$ classes & 25.9502 \\
\hline
{Standardization + U-Net Predict Noise} & 10 & 250 & $100 \times 10$ classes & 225.4544 \\
\hline
{Standardization + U-Net Predict Vector (Real Result)} & 10 & 250 & $100 \times 10$ classes & 112.7405 \\
\hline
{Each classes was split into two nearly equal parts.} & - & 250 & -& 3.2482 \\
\hline
\end{tabular}
\end{center}
\end{table*}

In the training process of diffusion models, in addition to the conventional parameters such as batch size, epochs, and learning rate, there are three important hyper-parameters that significantly affect the performance of the diffusion model.

One is the number of Diffusion steps.The number of diffusion steps represents how many steps of reverse denoising are needed to recover the original data, thus affecting the quality of the generated samples and the model's stability. A vector with only 34 dimensions is much smaller than the dimension of image data, so this paper selects 10 as the number of diffusion steps. After 10 reverse denoising steps, high-quality data can already be obtained. Moreover, as the number of diffusion steps increases, the quality of the generated data may even decrease. Additionally, a larger number of diffusion steps also increases the generation time of the data.

Another two types of hyper-parameters is the method of normalization and the output of U-Net network. We considered three types of methods of scaling data, [-1,1] normalization, [0,1] normalization and standardization.And two types of output that U-Net network generates, noise and real result. We randomly combine these different methods of two hyper-parameters. Then, we use generated data to calculate the average FID score over 10 classes for each combination. The calculation of FID is as shown in Equation \ref{fid}:
\begin{equation}
\label{fid}
\text{FID} = \|\mu_1 - \mu_2\|^2 + \text{Tr}(\Sigma_1 + \Sigma_2 - 2(\Sigma_1 \Sigma_2)^{\frac{1}{2}})
\end{equation}
Actually, this average FID score is obtained by calculating the FID 100 times and then averaging, which avoids randomness. Additionally, in order to prevent data range differences caused by different scaling methods, we restore the generated data to the original data range before calculating the FID. To demonstrate the quality of the generated data, we also divide the original data into two parts and calculate the FID between these two parts of real data. The result FID score is 3.2482. 

As can be seen from Table \ref{table_comparison}, the combination of [0,1] normalization and U-net predict real result has the nearest FID score to the real FID score. Although there is a gap of 1, the FID score is calculated after the data range has been expanded to the original data range, which significantly increases the FID score. In fact, if the FID is calculated directly using the normalized data, the FID score will show almost no difference from the real data.

\subsection{Experiments of Transformer Model}

During the training of Transformer, we take the training method of using weighted loss function and MSE loss function alternatively. Because the selection of weight isn't accurate, it may induce the former time steps have lower loss and the later time steps have higher loss. To smooth the curve line of loss, we use MSE loss function to train the model following the use of weighted loss function. This training procedure can make the loss at every time steps decrease normally. Although the loss at later time steps might be higher, it will not have a significant impact on sign language recognition.Because there are fewer meaningful actions at later time steps. Another small method we have taken is adding a random noise to the first time-steps. This noise conform to
0 mean and 0.1 variation.
\begin{equation}
    \epsilon \sim \mathcal{N}(0, 0.1)
\end{equation}
We are inspired by the theory of VAE. Through adding a random noise to the input, we can get a more robust and more universal model.

In this paper, We trained the model for 200 epochs with a learning rate of \(1 \times 10^{-4}\) and a weighted loss function. Then, we used the MSE loss function to smooth the loss for 100 epochs. Finally, we adjusted the learning rate to \(1 \times 10^{-5}\) and trained the model for 100 epochs using the weighted loss function. As for the selection of window size, we tried three different window size, 1, 3, 5. The comparison of different result will be shown in next subsection.

\begin{algorithm}[H]
\caption{Transformer Training Process}
\begin{algorithmic}[1]
    \Require Input data $X \in \mathbb{R}^{batch\_size \times 610 \times 34}$
    \Ensure Trained Transformer model
    
    \Statex \textbf{Prepare Input and Target:}
    \Statex $Encoder \gets X[:, 0:609, :]$ 
    \Statex $Decoder \gets X[:, 0:609, :]$ 
    \Statex $Label \gets X[:, 1:610, :]$ 
    
    \Statex \textbf{Forward Pass:}
    \Statex $\hat{Y} \gets \text{Transformer}(Encoder, Decoder)$ 
    
    \Statex \textbf{Loss Computation:}
    \Statex $\mathcal{L} \gets \text{Criterion}(\hat{Y}, Label)$ 
    
    \Statex \textbf{Backpropagation:}
    \Statex $\text{Update weights using } \nabla_\theta \mathcal{L}$
    
    \Statex \textbf{Repeat:} Steps 1–4 for each batch until convergence
    
\end{algorithmic}
\end{algorithm}

Finally, due to the lack of methods that can evaluate the quality of generated data, in this paper, we decided to use an indirect way to evaluate the quality of the results-by measuring the improvement of using augmented data compared to not using it.

\subsection{Results of Augmented Data}

\begin{figure}[!t]
    \centering
    \subfloat[window-size=1]{
        \includegraphics[width=0.3\textwidth]{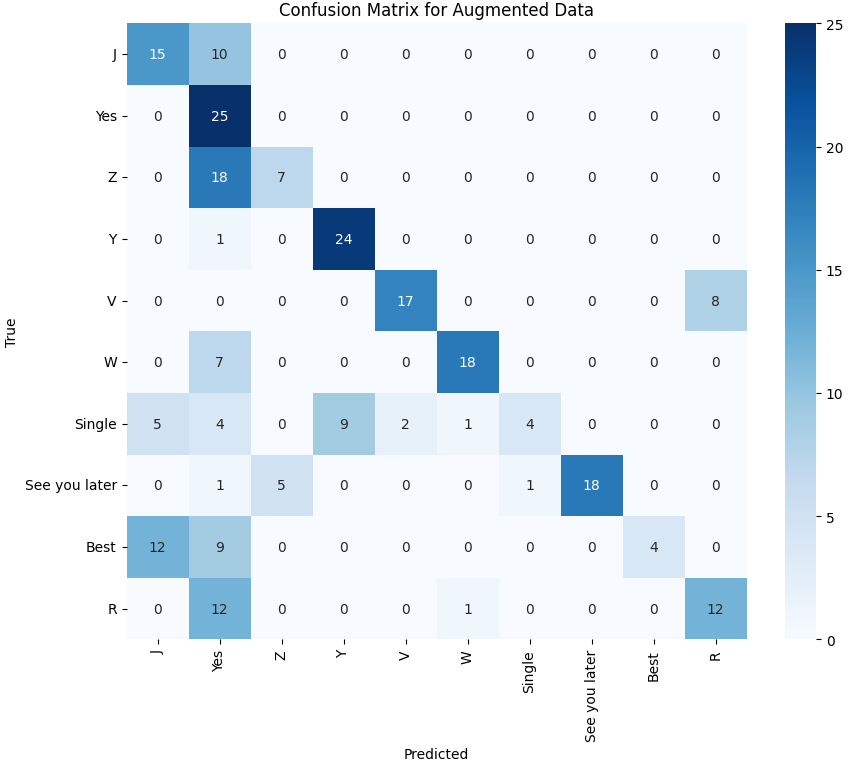} 
        \label{window-size=1}
    }
    \hfil
    \subfloat[window-size=3]{
        \includegraphics[width=0.3\textwidth]{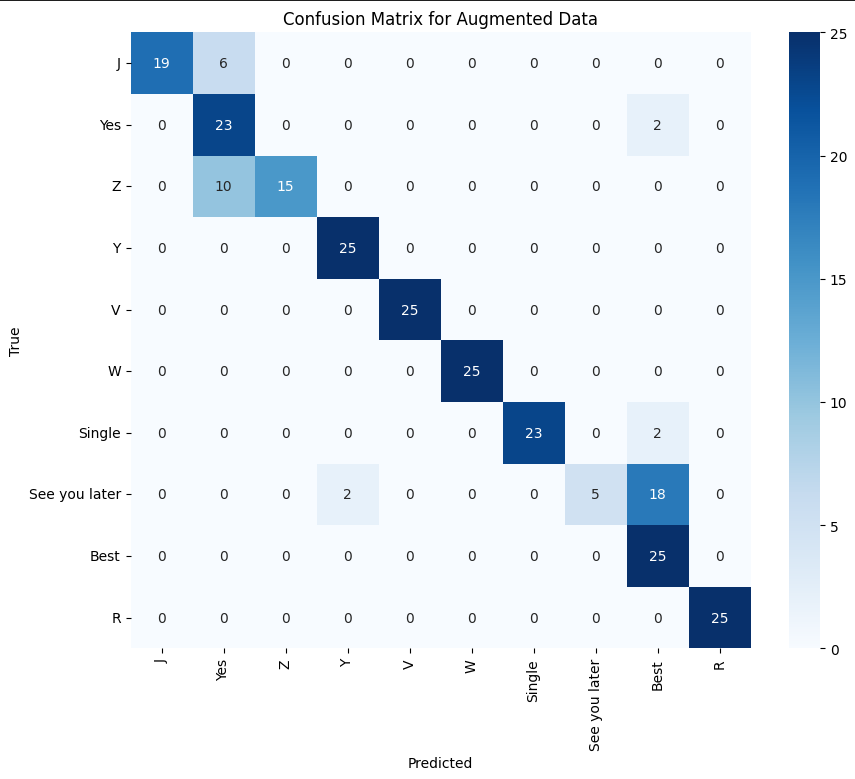} 
        \label{window-size=3}
    }
    \hfil
    \subfloat[window-size=5]{
        \includegraphics[width=0.3\textwidth]{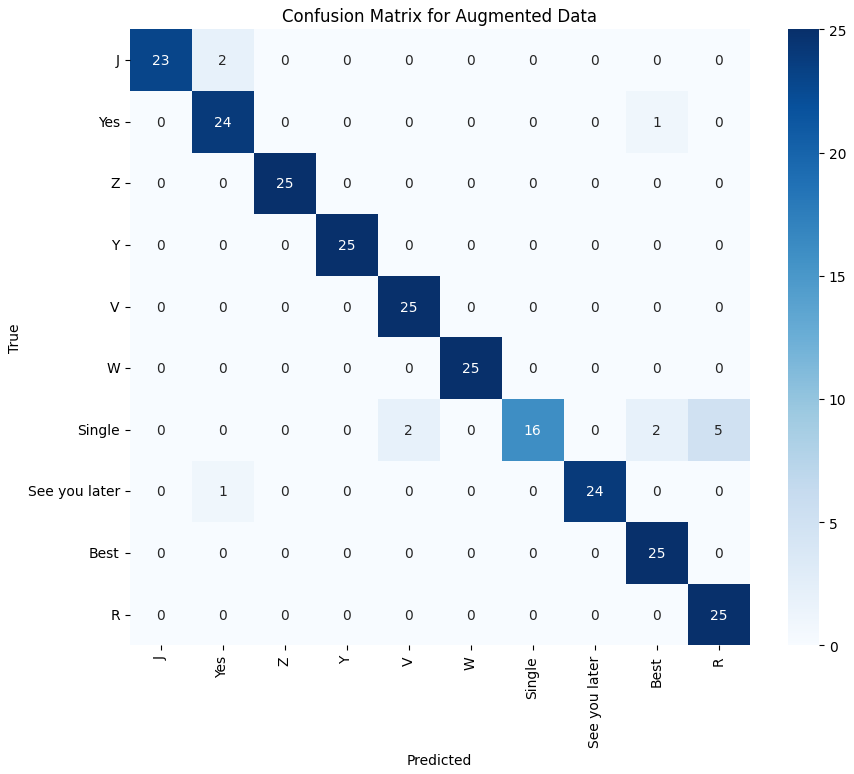} 
        \label{window-size=5}
    }
    \caption{Classification Results of Generated Data Across Different Window Sizes.}
    \label{classification}
\end{figure}

After training with window sizes of 1, 3, and 5, optimal parameters were used to generate 250 samples (25 per category) to augment the training set and assess classification performance improvements. The study employed CNN and Transformer models, but first verified that the generated data resembled real data. Due to the lack of standardized metrics for time-series data, an indirect evaluation was performed by having a classifier trained on real data classify the generated samples. Although this approach is only approximate—and misclassifications might indicate that generated samples extend beyond the real data distribution—it serves as a useful assessment tool. The classification model, trained on both the training and testing datasets, was then used to classify the generated data, with results shown in Figures \ref{classification}.

Figures show that the classification model achieved 57.6\%, 84\%, and 94.8\% accuracy for generated data using window sizes of 1, 3, and 5, respectively. This contradicts our initial hypothesis, which predicted optimal performance at window size 1 based on a presumed Markov decision process. However, confusion matrices reveal that a window size of 5 produces data that more closely resembles real data, leading to higher accuracy.

Next, we evaluated the impact of data augmentation. First, the model was trained on the original training set without augmentation and tested on the test set. To isolate the effect of data augmentation, we also generated 250 new samples (using noise addition and waveform distortion/restoration) to match the sample size of the augmented datasets. Finally, we combined the original training set with both the traditionally augmented data and the three deep neural network-generated augmented datasets to train the model.

The classification accuracy without augmented data was 57.6\%, indicating significant variability in the execution of the same action by different subjects, which hindered the model's ability to accurately distinguish actions within the same category. After augmenting the data using traditional methods, the classification accuracy improved to 77.2\%, an increase of 19.6\%. When combined with data augmented through deep neural network models, the classification accuracies improved to 75.2\%, 87.6\%, and 87.2\%, representing increases of 17.6\%, 30\%, and 29.6\%, respectively, compared to the baseline without augmentation. Compared to the data augmented using traditional methods, the accuracy increased by -2\%, 10.4\%, and 10\%. The results are summarized in Table \ref{accuracy_improvement} and Table \ref{increase_traditional_methods}.

\begin{table}[ht]
\begin{center}
\caption{Classification Accuracy Improvement over Original Data} 
\label{accuracy_improvement} 
\begin{tabular}{| c | c | c |}
\hline
\textbf{Method} & \textbf{Accuracy (\%)} & \textbf{Increase (\%)} \\ \hline
Without Augmented Data & 57.6 & - \\ \hline
With Traditional Augmented Data & 77.2 & +19.6 \\ \hline
window-size=1 Augmented Data (1) & 75.2 & +17.6 \\ \hline
window-size=3 Augmented Data (2) & 87.6 & +30.0 \\ \hline
window-size=5 Augmented Data (3) & 87.2 & +29.6 \\ \hline
\end{tabular}
\end{center}
\end{table}

\begin{table}[ht]
\begin{center}
\caption{Increase over Traditional Methods Augmented Dataset} 
\label{increase_traditional_methods} 
\begin{tabular}{| c | c | c |}
\hline
\textbf{Method} & \textbf{Accuracy (\%)} & \textbf{Increase (\%)} \\ \hline
With Traditional Augmented Data & 77.2 & - \\ \hline
window-size=1 Augmented Data (1) & 75.2 & -2.0 \\ \hline
window-size=3 Augmented Data (2) & 87.6 & +10.4 \\ \hline
window-size=5 Augmented Data (3) & 87.2 & +10.0 \\ \hline
\end{tabular}
\end{center}
\end{table}

These results highlight that using a window size of 3 or 5 leads to substantial improvements in the model's classification performance. While the augmented data generated with a window size of 3 showed lower similarity to the real data compared to a window size of 5, its greater contribution to performance improvement suggests that it encompassed more out-of-distribution data, effectively enhancing the model's robustness.

\section{Conclusion}

In this paper, we proposed a time-series data generation method that combined Diffusion model and Transformer model together. Through View Mask component, we made the Transformer pay more attention to some specific positions and used a new weighted loss function to make the waveform of the data change normally. This simple and straightforward methods exhibited effectiveness and rationality. But actually this methods didn't blend the Diffusion and Transformer together. They still belongs to two different, individual parts. In the future, we decided to try some methods that can blend Diffusion and Transformer together to make the overall model  become an end-to-end system. And find a loss function that can satisfy more practical application. No need to change loss function with the variation of model application.Make the model more universal and versatile. Most importantly, it is crucial to find a criterion that helps the model quantitatively demonstrate the authenticity and effectiveness of the generated data.

\section*{Acknowledgments}
This work was supported by JSPS KAKENHI Grant Number 22K12114 and NEDO Intensive Support for Young Promising Researchers Number 21502121-0 and JKA and its promotion funds from KEIRIN RACE.
 
\bibliographystyle{IEEEtran}
\bibliography{main}
\end{document}